\documentclass[11pt]{article}
\usepackage{nodalida2025}
\usepackage{times}
\usepackage{url}
\usepackage{latexsym}

\usepackage{url}
\usepackage{booktabs}       
\usepackage{amsfonts}       
\usepackage{nicefrac}       
\usepackage{microtype}      
\usepackage{xcolor}         
\usepackage{colortbl}
\usepackage{todonotes}
\usepackage{graphicx}
\usepackage{subcaption}
\usepackage{fontawesome5}

\usepackage{hyperref}
\usepackage[noabbrev,capitalize,nameinlink]{cleveref}

\newcommand{\sprogmodel}{{\textsc{SnakModel}-7B\textsubscript{base}}}
\newcommand{\snakmodel}{{\textsc{SnakModel}-7B}\textsubscript{instruct}}
\newcommand{\snak}{\textsc{SnakModel}}
\newcommand{\llama}{{\textsc{Llama2}-7B}}
\newcommand{\llamabase}{{\textsc{Llama2}-7B\textsubscript{base}}}
\newcommand{\llamachat}{{\textsc{Llama2}-7B\textsubscript{chat}}}

\newcommand{\itda}{INST\textsubscript{da}}
\newcommand{\cb}[1]{\color{blue}{\textbf{#1}}}
\newcommand{\co}[1]{\color{orange}{\textbf{#1}}}
\newcommand\blfootnote[1]{%
  \begin{NoHyper}
  \begingroup
  \hspace{-1.5em}
  \renewcommand\thefootnote{}\footnote{#1}%
  \addtocounter{footnote}{-1}%
  \endgroup
  \end{NoHyper}
}
\newcommand*\circled[1]{\tikz[baseline=(char.base)]{
            \node[shape=circle,draw,inner sep=.6pt] (char) {#1};}}
\newcommand{\eqcontrib}{\hspace{.3em}\raisebox{.07em}{\resizebox{1em}{!}{\circled{\tiny\faEquals}}}}

\aclfinalcopy 

\title{\includegraphics[scale=0.25]{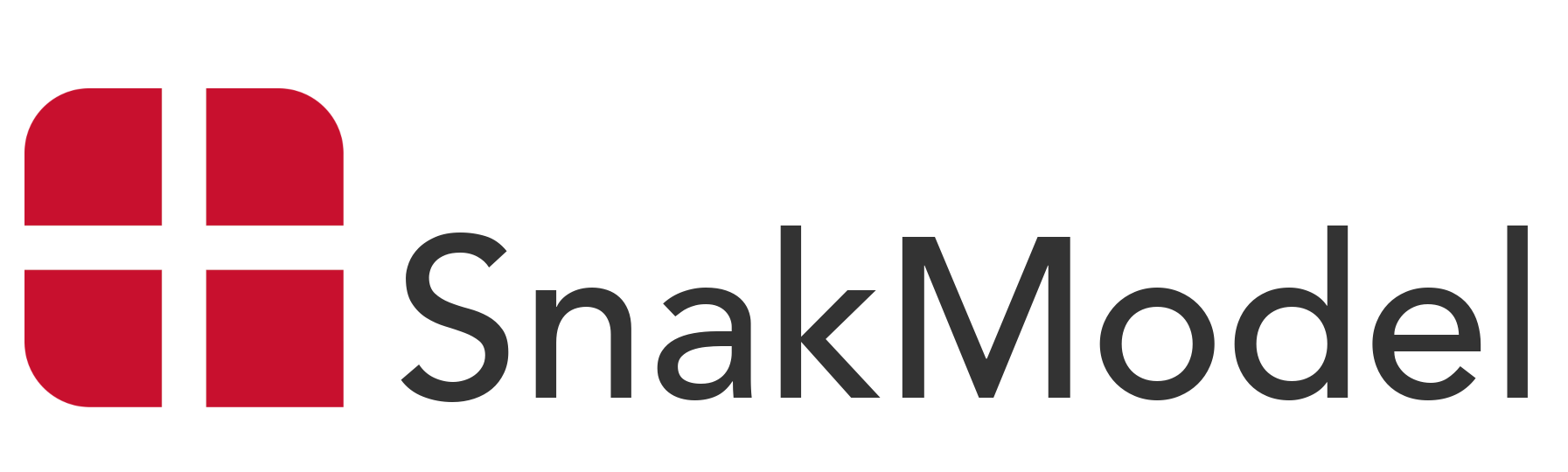}\\ Lessons Learned from Training an Open Danish Large Language Model}

\author{
Mike Zhang\textsuperscript{\eqcontrib{}\faWater\faRobot}\hspace{.4em}
Max Müller-Eberstein\textsuperscript{\eqcontrib{}\faCompass\faRobot}\hspace{.4em}
Elisa Bassignana\textsuperscript{\faCompass\faRobot}\hspace{.4em}
Rob van der Goot\textsuperscript{\faCompass\faRobot} \\
\textsuperscript{\faWater}Aalborg University, Denmark \\
\textsuperscript{\faCompass}IT University of Copenhagen, Denmark \\
\textsuperscript{\faRobot}Pioneer Center for Artificial Intelligence, Denmark \\
{\tt jjz@cs.aau.dk}\hspace{.9em}
{\tt \{mamy, elba, robv\}@itu.dk}
}

\date{}

\begin{document}
\maketitle
\blfootnote{\textsuperscript{\hspace{-1.5em}\eqcontrib{}\hspace{-.3em}} These authors contributed equally.}
\begin{abstract}
\looseness=-1
We present \snak{}, a Danish large language model (LLM) based on \llama, which we continuously pre-train on 13.6B Danish words, and further tune on 3.7M Danish instructions. As best practices for creating LLMs for smaller language communities have yet to be established, we examine the effects of early modeling and training decisions on downstream performance throughout the entire training pipeline, including (1) the creation of a strictly curated corpus of Danish text from diverse sources; (2) the language modeling and instruction tuning training process itself, including the analysis of intermediate training dynamics, and ablations across different hyperparameters; (3) an evaluation on eight language and culturally-specific tasks. Across these experiments \snak{} achieves the highest overall performance, outperforming multiple contemporary \llama{}-based models. By making \snak{}, the majority of our pre-training corpus, and the associated code available under open licenses, we hope to foster further research and development in Danish Natural Language Processing, and establish training guidelines for languages with similar resource constraints.\footnote{The code and data scripts are available here:\\\url{https://github.com/nlpnorth/snakmodel/}.}
\end{abstract}

\section{Introduction}
The landscape of large language models (LLMs) has seen rapid expansion, with an increasing trend towards open-weight releases for a broader range of languages. Notable English-centric examples include Pythia~\cite{biderman2023pythia}, Vicuna~\cite{zheng2023judging}, Mistral~\cite{jiang2023mistral}, Qwen~\cite{bai2023qwen}, Llama2~\cite{touvron2023llama}, Llama3~\cite{dubey2024llama}, OLMo~\cite{groeneveld-etal-2024-olmo}, and Phi~\cite{abdin2024phi}. Simultaneously, recent efforts have extended LLMs to multilingual settings, including models such as mT5~\cite{xue2020mt5}, Bloom~\cite{le2023bloom}, Aya~\cite{ustun-etal-2024-aya,singh-etal-2024-aya}, RomanSetu~\cite{j-etal-2024-romansetu}, and EuroLLM~\cite{martins2024eurollmmultilinguallanguagemodels}.

As anglocentric and/or multilingual LLMs have nonetheless continued struggling to adapt to lower-resource settings---especially with respect to pragmatic and sociolinguistic factors \citep{hershcovich-etal-2022-challenges,cao-etal-2023-assessing,naous-etal-2024-beer,wang-etal-2024-countries}---there is growing interest in language-specific LLMs, either tailored to a single language (see Related Work;~\cref{sec:relatedWork})
or specialized for a small set of similar languages~\cite{viking2024, dou2024sailor}. 
However, the best practices for creating such language-adapted LLMs have yet to be established---especially for smaller language communities with resource limitations with respect to data, compute, or both.

Danish offers a particularly interesting testbed among these smaller languages. As a mid-resource language, which is typologically related to English and has largely overlapping character sets, it has sufficient textual data for LLM adaptation, yet is far from the levels of its neighbors (e.g., Swedish; \citealp{ekgren-etal-2024-gpt}). Additionally, it lacks advanced resources like native instruction-tuning data or human-preference data, making it necessary to use translated datasets for which the downstream effects on model functionality are not yet well understood.
Linguistically, Danish has also been shown to be more challenging to learn for humans than its neighbors due its phonological complexity ~\cite{DanishAsAWindow, DanishPuzzle}, which results in downstream effects on discourse, such as additional conversational redundancy \citep{DanishPuzzle,Language-SpecificConstraints}.

With the goal to provide the Danish community with a custom-adapted resource, as well as to establish better-grounded guidelines for creating LLMs in languages with similar linguistic characteristics and resource constraints, we present and analyze \textsc{SnakModel}-7B\textsubscript{base/instruct}, two LLMs designed specifically for the Danish language. Our base model builds upon \llama{}, which we continuously pre-train on a diverse collection of Danish corpora comprising 350M documents (sentences/paragraphs) and 13.6B words, before tuning it on 3.7M Danish instruction-answer pairs. We evaluate our model against contemporary \llama{}-based models on the Danish part of the ScandEval benchmark \citep{nielsen-2023-scandeval} that encompasses both language and culture-specific tasks. By releasing not just the related artifacts (final model, intermediate checkpoints, pre-training data, code), but by also analyzing the effects of early decisions in the training and model design process on intermediate training dynamics and downstream performance, we aim to provide resources that are not just relevant for Danish, but for LLM adaptation in general.

\paragraph{Contributions.} This work contributes:

\begin{itemize}
\itemsep0em
    \item A large, diverse, high-quality collection of Danish corpora, totaling 350M documents with 13.6B words (\cref{sec:data}). We provide scripts to collect and process the data.
    \item \textsc{SnakModel-7B}\textsubscript{base/instruct}, two open-weight 7B-parameter language models continuously pre-trained and instruction-tuned specifically for Danish, for which we release all related artefacts, and extensively analyze the model's intermediate training dynamics (\cref{sec:training}).
    \item An evaluation comparing \snakmodel{} and contemporary Danish models, which analyzes performance with respect to language and cultural tasks (\cref{sec:results}).
    \item A consolidation of our findings into recommendations for efficiently training LLMs under similar resource constraints (\cref{sec:guidance}).
\end{itemize}

\section{Related Work}
\label{sec:relatedWork}

\paragraph{Continuously Pre-trained LLMs.}
Previous work has shown that for both encoder and decoder language models (LM), continuous pre-training is the de facto standard for adapting an LM to a specific domain~\cite{han-eisenstein-2019-unsupervised, alsentzer-etal-2019-publicly, lee2020biobert, gururangan-etal-2020-dont, nguyen-etal-2020-bertweet} or another language, such as German~\cite{leo2023lm}, Spanish and Catalan~\cite{aguila2023}, Finnish~\cite{luukkonen-etal-2023-fingpt}, Dutch~\cite{rijgersberg2023geitje, vanroy2024fietje}, Italian~\cite{bacciu-etal-2024-dantellm}, Japanese~\cite{group2024rakutenai}, Basque~\cite{etxaniz-etal-2024-latxa}, Swedish~\cite{aisweden2024llama}, Modern Greek~\cite{voukoutis2024meltemiopenlargelanguage}, Norwegian~\cite{nora2024llm}, or multiple languages~\cite{xue2020mt5, alves2024tower, ustun-etal-2024-aya, costa2022no, martins2024eurollmmultilinguallanguagemodels, dou2024sailor, nguyen-etal-2024-seallms, aryabumi2024aya, dang2024aya}. 

\paragraph{Open Large Language Models.}
Recent open language models can be broadly divided into \textit{open-source} LLMs and \textit{open-weight} LLMs. The main difference is that open-weight releases include at least a basic description of the training data, as well as the model weights themselves. For open-source LLMs, instead, the (non-trivial) expectation is to have all resources released, including data, training scripts, evaluation scripts, and model weights. We follow previous endeavors such as Pythia~\cite{biderman2023pythia}, OLMo~\cite{groeneveld-etal-2024-olmo}, Latxa~\cite{etxaniz-etal-2024-latxa}, and Meltemi~\cite{voukoutis2024meltemiopenlargelanguage}, and release most sources of our training data, including training and evaluation scripts, as well as the model weights.

\paragraph{Danish Language Resources.}
In-language resources are the fundamental building block for further training an LLM for the Danish language. There are several open-source toolkits for Danish, including models and datasets~\cite{pauli-etal-2021-danlp, enevoldsen2021dacy}. Additionally, there are several Danish-specific large corpora of raw text, such as DaNewsroom~\cite{varab-schluter-2020-danewsroom} and Danish Gigaword~\cite{stromberg-derczynski-etal-2021-danish}. Additionally, Danish subsets can be found in public resources built on crawled web data such as CommonCrawl~\cite{wenzek-etal-2020-ccnet} and CulturaX~\cite{nguyen2023culturax}. In this work, we collect and combine a variety of sources for wider coverage, before pre-processing them through a joint pipeline.

\paragraph{Danish Large Language Models.} 
Previous endeavors at training LLMs that cover the Danish language include \citet{ciosici2022trainingt5usinglabsized}, who trained a T5 model~\cite{t5} for Danish.
More recently, within the decoder-only family of models, Munin~\cite{dfm2024munin} and Viking~\cite{viking2024} were released.
Munin is based on Mistral-7B (v0.1~\citealp{jiang2023mistral}) and is further pre-trained on the Danish Gigaword Corpus~\cite{stromberg-derczynski-etal-2021-danish} containing 1B words. However, the model seems to underperform compared to its base model counterpart, indicating some form of catastrophic forgetting.
Viking is based on \llama{}, and pre-trained from scratch on a mix of English, Finnish, Swedish, Danish, Norwegian, Icelandic and code~\cite{viking2024}.
In this work, \snakmodel{} is continuously pre-trained for Danish, and outperforms its original checkpoint, as well as all other currently available Danish models with a comparable size.

\section{Data \& Pre-processing}\label{sec:data}

\subsection{Pre-training}

Our Danish pre-training data, as shown in~\cref{tab:data}, initially encompassed 927M documents and 24.6B words, as measured by the Unix \texttt{wc} command. The data is sourced from diverse platforms, for which we verify appropriate licensing (wherever possible), and include:

\paragraph{Bookshop (\texttt{cc-by-4.0}).} EU Bookshop text from OPUS~\citep{tiedemann-2012-parallel}, as integrated by~\citet{skadins-etal-2014-billions}. It contains well-edited, official EU publications across diverse topics, converted automatically from PDFs.

\paragraph{CC-100 (\texttt{UNK}).} A cleaned version of a 2018 CommonCrawl dump~\citep{wenzek-etal-2020-ccnet}, reproducing data from~\citet{conneau-etal-2020-unsupervised}. It consists of web data, filtered using the fastText language classifier~\citep{joulin-etal-2017-bag}.

\paragraph{CulturaX (\texttt{odc-by + cc0}).} mC4 (v3.1.0) combined with accessible OSCAR corpora~\citep{nguyen2023culturax}.

\paragraph{DaNewsroom (\texttt{UNK}).} Scraped from 19 news outlets~\citep{varab-schluter-2020-danewsroom}, originally for summarization. We use the full news articles instead of summaries.

\paragraph{Dawiki (\texttt{cc-by-sa}).} Cleaned Wikipedia data from 01-01-2024~\citep{Wikiextractor2015}.

\paragraph{FTSpeech (\texttt{FT-OD + FT-TV}).} A transcription-based corpus from Danish parliamentary data~\citep{kirkedal2020ft}, used in language modeling due to its large text volume.\footnote{FT-OD and FT-TV refer to Folketing’s open data and Folketing TV license.}

\paragraph{Gigaword (\texttt{cc0 + cc-by}).} Danish Gigaword~\citep{stromberg-derczynski-etal-2021-danish} covers a range of domains including wiki, books, web, and social media data.

\paragraph{OpenSubtitles (\texttt{UNK}).} Danish data from OPUS OpenSubtitles~\citep{lison-tiedemann-2016-opensubtitles2016,tiedemann-2016-finding}.\footnote{\url{http://www.opensubtitles.org/}}

\paragraph{Reddit (\texttt{UNK}).} Danish Reddit data from ConvoKit~\citep{chang-etal-2020-convokit}, specifically \texttt{Denmark.corpus.zip}.

\paragraph{Twitter (\texttt{MIT}).} Data from the public Twitter stream,\footnote{\url{https://archive.org/details/twitterstream}} reclassified using our own pipeline due to inaccurate language labels.

\begin{table}[t]
    \centering
    \small
    \begin{tabular}{l rr c rr}
    \toprule
    \textsc{\textbf{Dataset}}    & \multicolumn{2}{c}{\textsc{\textbf{Original}}} && \multicolumn{2}{c}{\textsc{+ \textbf{FastText}}}   \\
                             \cline{2-3}                  \cline{5-6}                    
                             & Docs & Words          && Docs  &  Words                        \\
    \midrule
    Bookshop                 & 8.65M   & 208M              && 6.80M    & 187M                 \\
    CC-100                   & 344M    & 7.82B             && 256M    & 7.16B                 \\
    CulturaX                 & 449M    & 14.8B             && 333M    & 13.7B                 \\
    DaNewsroom               & 24.2M   & 391M              && 11.3M   & 369M                  \\
    Dawiki                   & 1.70M   & 62.4M             && 1.20M   & 57.3M                 \\
    FTSpeech                 & 2.03M   & 43.3M             && 1.69M   & 40.9M                 \\
    Gigaword                 & 62.0M   & 1.02B             && 39.3M   & 898M                  \\
    OpenSubtitles            & 30.2M   & 207M              && 19.6M   & 156M                  \\
    Reddit                   & 4.50M   & 73.9M             && 2.37M   & 64.0M                 \\
    Twitter                  & 1.69M   & 21.9M             && 406K    & 6.61M                 \\
    \midrule
    \textbf{\textsc{Total}}           & 927M    & 24.6B             && 672M    & 22.6B                 \\
    \multicolumn{2}{c}{\hspace{1em} + \textsc{\textbf{Deduplication}}} &&& \textbf{350M} & \textbf{13.6B}\\
    \bottomrule
    \end{tabular}
    \caption{\textbf{Preprocessing Steps.} Data in number of words using \texttt{wc} command. In the \textbf{Original} column, we already use a pre-defined Danish slice of the dataset. In the \textbf{FastText} column, we apply another round of language identification to the data. In the \textbf{Deduplication} row, we combine all data and deduplicate it, which results in around 350M documents and 13.6B words for the pre-training process.}
    \label{tab:data}
\end{table}

\paragraph{} To refine the overall concatenated dataset, we implemented a preprocessing pipeline using \texttt{fastText}~\citep{joulin-etal-2017-bag}\footnote{Using the \texttt{lid.176.bin} model with a threshold of 0.6.} for language identification and \texttt{text-dedup}~\citep{chenghao_mou_2023_8364980}\footnote{\url{https://github.com/ChenghaoMou/text-dedup}} for text deduplication. The language identification process eliminated 28\% of the documents while retaining 92\% of the tokens, indicating that many short documents were removed, where language prediction was less confident. The deduplication step further reduced the corpus by 48\% in document count and 40\% in token count. 
We anticipated significant content overlap between CC-100 and CulturaX, which underlines the importance of deduplication in creating a more efficient and representative dataset. These preprocessing steps reduced our dataset to approximately 350M documents with 13.6B words. Following the open LLM approach, we release all scripts used for collecting and processing the data. 

\subsection{Instruction Tuning}\label{sec:instruction-datasets}

As for most mid-to-low resource languages, Danish~\cite{joshi-etal-2020-state} currently lacks human-generated instruction tuning data, and instead relies on automatically translated data from English, which itself may be generated by LLMs. From these sources, we select the following three after manually inspecting them for quality:

\paragraph{SkoleGPT \citep{skolegpt}}: A subset of OpenOrca \cite{OpenOrca}, which was automatically translated into Danish and filtered for quality, containing 21.6k instruction-output pairs.

\paragraph{Danish OpenHermes \citep{danishopenhermes}}: A subset of the automatically generated OpenHermes dataset \cite{OpenHermes}, which was automatically translated into Danish. It contains 98.7k instruction-output pairs.

\paragraph{Aya Collection \citep{singh-etal-2024-aya}}: A collection of 44 datasets, which were automatically translated based on instruction templates from fluent speakers. While the underlying Aya Dataset, on which these translations are based, was created by native speakers, the Danish portion of this data contains less than 100 instances, leading us to opt for the translations instead. We use 3.6M instruction-output pairs from the Danish subset of the data.

\paragraph{} Together, these data sources sum up to a total of 3.7M instruction-answer pairs, which we train \sprogmodel{} on in \cref{sec:instruction-tuning}.

\subsection{Evaluation Framework}\label{subsec:evaluation}
For evaluation, we use the \textsc{ScandEval} benchmark~\citep{nielsen-2023-scandeval} covering eight tasks. The tasks cover named entity recognition (\textsc{NER}; DANSK by~\citealp{hvingelby-etal-2020-dane}), sentiment analysis (\textsc{Senti}; AngryTweets by~\citealp{pauli-etal-2021-danlp}), linguistic acceptability (\textsc{LA}; ScaLA\footnote{Based on the Universal Dependencies dataset from \cite{kromann2004danish}.}), abstractive summarization (\textsc{Summ}; Nordjylland-News by~\citealp{kinch2023nordjylland}), commonsense reasoning (\textsc{CSR}; translated HellaSwag by~\citealp{zellers-etal-2019-hellaswag}), and question answering (\textsc{QA}; ScandiQA\footnote{ScandiQA is a translation of the English MKQA dataset~\cite{longpre2021mkqa} and does not strictly focus on Scandinavian knowledge.}). The benchmark also include culture-specific datasets, namely Danske Talemåder (\textsc{TM};~\citealp{nielsen-2023-scandeval}), which prompts for meanings behind common proverbs, and a collection of official Danish Citizenship Tests (\textsc{CT};~\citealp{nielsen2024citizen}). Evaluation metrics differ per task, and are indicated as $F_1$, macro-averaged $F_1$ ($mF_1$), micro-averaged $F_1$ ($\mu F_1$), BERTScore (BERTS.;~\citealp{Zhang2020BERTScore}), and Accuracy (Acc.).

\section{Model Training}\label{sec:training}

\subsection{Language Modeling Pre-training}

\begin{table}[t]
\centering
\small
\begin{tabular}{ll}
\toprule
\textbf{Parameter} & \textbf{Value} \\
\midrule
\multicolumn{2}{c}{\textit{Data Split}} \\
\midrule
Training data & 96.9\% \\
Validation data & 3.1\% \\
\midrule
\multicolumn{2}{c}{\textit{Training Configuration}} \\
\midrule
Vocabulary size & 32,000 \\
Context length & 4,096 \\
Training steps & 12,500 \\
Warmup steps & 1,250 \\
Number of epochs & 1 \\
Global batch size & 512 \\
\midrule
\multicolumn{2}{c}{\textit{Optimizer Parameters (AdamW)}} \\
\midrule
$\beta_1$; $\beta_2$ & 0.9; 0.95 \\
$\epsilon$ & $10^{-5}$ \\
Peak learning rate & $1.5 \times 10^{-5}$ \\
Minimum learning rate & $5 \times 10^{-8}$ \\
Weight decay & 0.1 \\
Gradient clipping & 1.0 \\
\bottomrule
\end{tabular}
\caption{\textbf{Pre-training Hyperparameters and Configuration Details.} We show the hyperparameter details of \sprogmodel{} pre-training.}
\label{tab:training-params}
\end{table}

\paragraph{Training Details.} We continuously pre-train from \llamabase{}~\cite{touvron2023llama}. We show configuration and hyperparameter details in~\cref{tab:training-params}. For further pre-training and fine-tuning, we make use of the Megatron-LLM library~\citep{epfmgtrn}, based on the Megatron-LM library.\footnote{\url{https://github.com/NVIDIA/Megatron-LM}.} We use the same tokenizer as \llama{}, byte-pair encoding (BPE;~\citealp{sennrich-etal-2016-neural}) as implemented in the SentencePiece toolkit~\citep{kudo-richardson-2018-sentencepiece}, with a vocabulary size of 32K subwords. As Danish and English share the same Indo-European language family, we assume large overlap in vocabulary subwords. Hence, we do not re-train nor extend the vocabulary.

\paragraph{Hardware and Emissions.} \sprogmodel{} is trained on private infrastructure with one node, containing four NVIDIA A100-PCIe 40GB GPUs. The node has an AMD Epyc 7662 128 Core Processor and 1TB of RAM. Total time of training took 8,928 GPU hours (93 days $\times$ 24 hours $\times$ 4 GPUs) between March--June 2024. The average carbon efficiency was 0.122 $kgCO_2eq/kWh$ during this time in Denmark.\footnote{According to \url{https://app.electricitymaps.com/map}.} This is equivalent to 272.3 $kg$ $CO_2$ $eq.$ emitted, based on the Machine Learning Impact calculator \citep{lacoste2019quantifying}.\footnote{\url{https://mlco2.github.io/impact}.}

\begin{figure}[t]
    \centering
    \includegraphics[width=.8\linewidth]{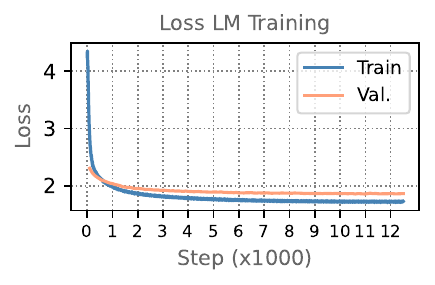}
    \caption{\textbf{\sprogmodel{} Pre-training Behaviour.} We report the stable language model loss during training and validation.}
    \label{fig:ptb}
\end{figure}

\paragraph{Loss Trajectories.}
In \cref{fig:ptb}, we show the continuous pre-training process of \sprogmodel{} in terms of loss curve based on perplexity. 
The loss shows a declining gain over time. We speculate that the model is close to convergence or that the learning rate is reduced, although previous work has shown that downstream performance can still increase with more training after loss and perplexity have converged~\cite{liu2023same}.

\paragraph{Leakage.} The training data of \llama{} is not public. However, since it was released in July 2023 after the ScandEval benchmark, we investigate potential test data leakage by prompting the model for information about the dataset (inspired by~\citealp{leak,balloccu-etal-2024-leak}), as well as completions for the first five sentences of each dataset. This process yielded no evidence that the evaluation datasets were included during training.

For \sprogmodel{}, we have access to all training data, such that we can search for 200 random 8-grams from each of our datasets in the raw data. We find that a small amount (6/200) of the tweets from AngryTweets are included in our Twitter sample (without labels). The DANSK NER dataset was completely included (without labels), as it was sampled from Gigaword, and many parts of the ScaLA dataset were also included in its original form in GigaWord and CC100. The code for all leakage tests is included in our code repository.

\subsection{Instruction Tuning}\label{sec:instruction-tuning}

Starting from \sprogmodel{}, we train our model on the Danish instruction datasets outlined in \cref{sec:instruction-datasets}.

\paragraph{Training Details.}
For instruction tuning, we opt for the more parameter-efficient low-rank adaptation (LoRA; \citealp{hu2022lora}), to enable faster iterations across multiple ablations (different template formats and base models), and to more easily analyze the intermediate training dynamics (\cref{sec:training-dynamics}). Nonetheless, we choose a substantially higher-parameter setup than is commonly employed when using LoRA \citep{hu2022lora,dettmers2023qlora}, in order to approximate full fine-tuning as closely as possible given our computational budget. Specifically, we use rank $r = 128$ adaptation matrices, which are applied to all parameters within the model without quantization \citep{dettmers2023qlora}. We train for one epoch over our instruction data using the AdamW optimizer with a constant learning rate of $2\times 10^{-4}$, and a global batch size of 64.

\begin{figure*}[t]
    \centering
    \begin{subfigure}[c]{.32\textwidth}
        \centering
        \includegraphics[width=\textwidth]{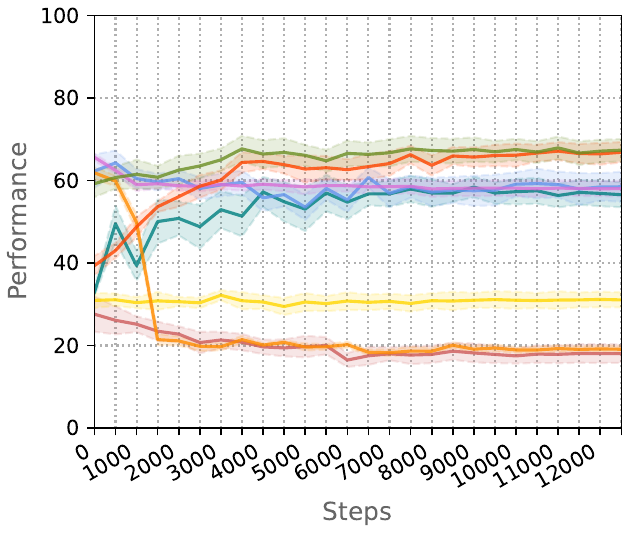}
        \caption{\sprogmodel{}}\label{fig:sprogmodel-training-dynamics}
    \end{subfigure}
    \begin{subfigure}[c]{.32\textwidth}
        \centering
        \includegraphics[width=\textwidth]{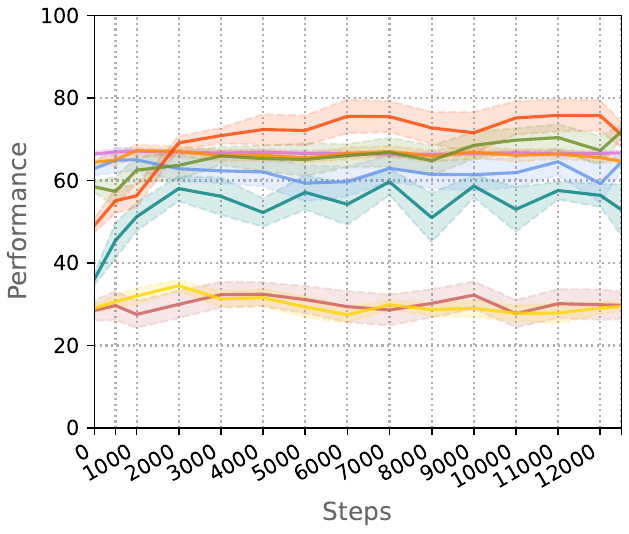}
        \caption{\snakmodel{}}\label{fig:snakmodel-training-dynamics}
    \end{subfigure}
    \begin{subfigure}[c]{.32\textwidth}
        \centering
        \includegraphics[width=\textwidth]{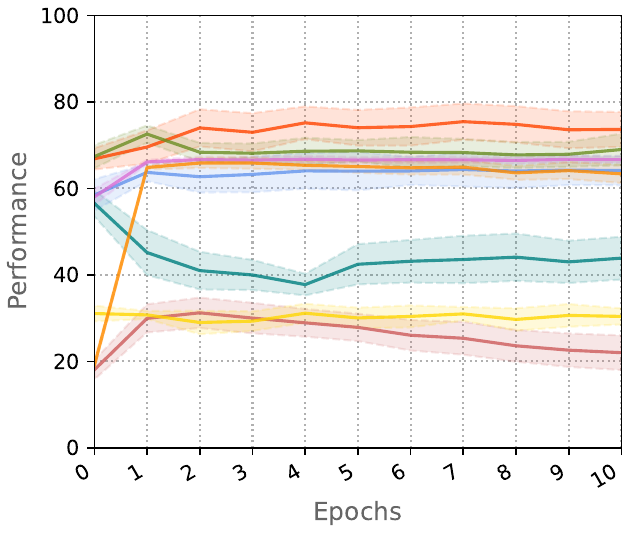}
        \caption{Instruction Tuning Dynamics}\label{fig:instruction-tuning-dynamics}
    \end{subfigure}\\[.5em]
    \begin{subfigure}[c]{.6\textwidth}
        \centering
        \includegraphics[width=\textwidth]{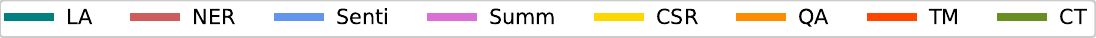}
    \end{subfigure}
    \caption{\textbf{\textsc{SnakModel} Training Dynamics} of LM pre-training, instruction tuning, and multi-epoch instruction tuning, as measured on the ScandEval (validation) tasks of linguistic acceptability (LA), named entity recognition (NER), sentiment analysis (\textsc{Senti}), summarization (\textsc{Summ}), commonsense reasoning (\textsc{CSR}), question answering (\textsc{QA}), proverb meaning (\textsc{TM}), and citizenship tests (CT).}
\end{figure*}

\paragraph{Instruction Template.}

The formatting of instruction-answer pairs is an important design decision with significant downstream impacts \citep{sclar2024quantifying}. For our adaptation context (\llama{} + Danish), we therefore ablate across three templates: (1) \textsc{Concat}, which concatenates instructions and answers; (2) \textsc{Chat}, which wraps the instruction in special \texttt{[INST]}/\texttt{[/INST]} delimiters following \llamachat{}\footnote{Note that these delimiters are not split by the tokenizer.}; (3) \textsc{Alpaca}, following a multi-line format with instruction/input/answer headers \citep{wang-etal-2023-self-instruct}, which we translate into Danish.

Instruction tuning using the \textsc{Chat} format leads to the highest overall scores on the validation split of our evaluation benchmark (56.37 avg.). \textsc{Concat} performs comparably (55.52 avg.), however we observe that models trained using this template frequently generate continuations to an instruction, instead of an answer.
\textsc{Alpaca} performs worst (53.26 avg.), and we observe that when prompting models without correctly terminating the instruction, the \textsc{Chat} model consistently terminates the instruction on its own (by generating \texttt{[/INST]}), while the \textsc{Alpaca} model often struggles to do so.

\subsection{Training Dynamics}\label{sec:training-dynamics}

We next investigate our models' intermediate training dynamics to establish how much language modeling and/or instruction tuning are required to obtain a certain level of performance (evaluated according to \cref{subsec:evaluation}), and whether these trajectories differ across task types.\footnote{The intermediate checkpoints can be found here: \url{https://huggingface.co/NLPnorth/snakmodel-7b-base/tree/main} for \sprogmodel{} and \url{https://huggingface.co/NLPnorth/snakmodel-7b-instruct} for \snakmodel{}.}

\paragraph{Language Modeling.}
By tracking the validation performance of the non-instruction-tuned \sprogmodel{} checkpoints across pre-training, we aim to identify when the English base model begins adapting to Danish. \cref{fig:sprogmodel-training-dynamics} shows performance on the Danish ScandEval tasks from start (\llamabase{}) to finish (\sprogmodel{}). For \textsc{Senti}, \textsc{Summ} and \textsc{CSR}, performance remains relatively consistent, while for \textsc{LA}, \textsc{TM} and \textsc{CT} performance gradually increases until 4,000--6,000 steps before converging.

Meanwhile, we see performance decreases for \textsc{NER} and \textsc{QA}, with the latter dropping from 61.9\% F1 to around 20\% within the first 2,000 steps. We attribute these changes to two respective hypotheses: for \textsc{NER}, answers are enforced to be in JSON-format in ScandEval. As our pre-training data consists exclusively of natural language, the model's output distribution may skew away from tokens such as ``\textsc{\{\}}'', required for this task. For QA, we qualitatively observe that \sprogmodel{} tends to generate continuations to the provided questions, instead of answers. Additionally, it does so in Danish, which may be detrimental to performance, since many answers in \textsc{QA} are English names.

\paragraph{Instruction Tuning.}
Next, we investigate the effect of applying instruction tuning at different points during Danish pre-training, in order to assess when it starts becoming beneficial. \cref{fig:snakmodel-training-dynamics} shows the validation performance of intermediate \sprogmodel{} checkpoints after instruction-tuning, i.e., from \llamabase{} + \itda{} (instruction-tuning on Danish instruction--completion pairs) until our final \snakmodel{} (fully pre-trained \sprogmodel{} + \itda{}). Once again, performance for most tasks is surprisingly stable throughout training. We further do not observe the same performance drops for \textsc{NER} and \textsc{QA} as during language modeling pre-training, showing that instruction tuning recovers these original functionalities. Additionally, we observe a general performance increase across the board. In particular, performance for \textsc{LA}, \textsc{TM}, and \textsc{CT} climbs and converges after 2,000--5,000 steps of Danish pre-training, and subsequent instruction-tuning. This indicates that training on less than half of our corpus may already be sufficient to obtain close-to-final performance. Interestingly, the largest performance improvements are observed for benchmark tasks based on Danish data, instead of translations (e.g., LA, TM, CT).

In terms of the training dynamics of instruction tuning itself, \cref{fig:instruction-tuning-dynamics} shows how one epoch of instruction tuning is already sufficient to obtain most performance gains, including the performance recovery of \textsc{NER} and \textsc{QA}. While there may be some benefit to one or two additional instruction tuning epochs, we believe that at this scale, they can be skipped in favor of efficiency. Since the use of duplicate data across epochs has however also been shown to negatively affect downstream performance \citep{biderman2023pythia}, we leave the exploration of this trade-off to future work.

\begin{figure}[t]
    \centering
    \includegraphics[width=\linewidth]{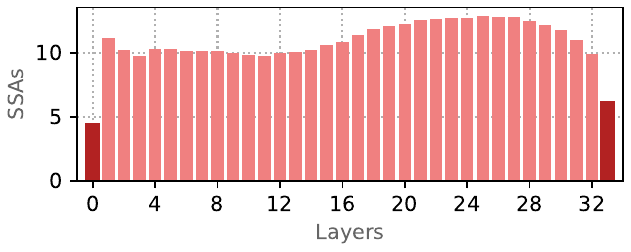}
    \caption{\textbf{Layer-wise Weight Divergence of \sprogmodel{}} as measured in total SSAs. Darker bars represent \textsc{Emb} and LMH respectively.}
    \label{fig:ssa-layers}
\end{figure}

\paragraph{Weight Divergence Analysis.}

Lastly, we take a closer look at changes \textit{within} the model to identify which parameters are most strongly affected by Danish language adaptation. To measure weight divergence, we follow \citet{me2024hypernetworks} and measure the principal subspace angles (SSAs; \citealp{knyazev2002ssa}) of each parameter before and after adaptatation (0$^{\circ}$/90$^{\circ}$ $\leftrightarrow$ similar/dissimilar). Across layers, \cref{fig:ssa-layers} shows how there is a slightly higher rate of change towards the penultimate layers of the model. This may be representative of cross-lingual encoding early in the model, and subsequent target language specialization in later layers \citep{wendler-etal-2024-llamas}.

\begin{figure}[t]
    \centering
    \includegraphics[width=\linewidth]{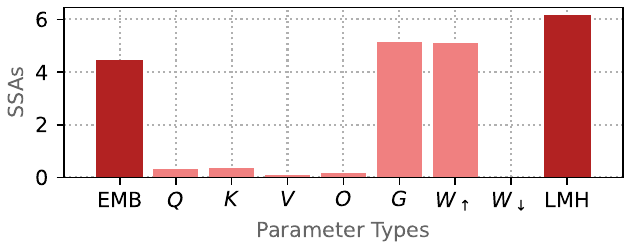}
    \caption{\textbf{Parameter-wise Weight Divergence of \sprogmodel{}} as measured in mean SSA. Darker bars represent \textsc{Emb} and LMH respectively.}
    \label{fig:ssa-parameters}
\end{figure}

\cref{fig:ssa-parameters} provides a more granular view of which parameter types are changing within each layer: Most updates per layer appear to be concentrated in the gate $G$ and up-projection $W_\uparrow$ of the SwiGLU feed-forward block \citep{shazeer2020glu}, while the down-projection $W_\downarrow$ and self-attention parameters ($Q$, $K$, $V$, $O$) are relatively unaffected. For the self-attention parameters, we hypothesize that this lack of change could be an effect of the relatively high syntactic similarity of English and Danish, requiring less adaptation for in-sequence dependencies. Interestingly, this pattern is also observed when adapting speech recognition models to under-resourced settings \citep{me2024hypernetworks}.

The initial embedding layer (\textsc{Emb}) as well as final language modeling head (\textsc{LMH}) also diverge to a comparable degree as $G$ and $W_\uparrow$, which is to be expected given their importance to receiving and generating text in a new language. In terms of token-level changes within \textsc{Emb} and \textsc{LMH} (as measured by the absolute difference of each token row before and after adaptation), we observe larger updates to subwords, which occur both in Danish and other Germanic languages (e.g., ``\_er'', ``\_ik'', ``\_billion''), while subwords in other scripts appear to be least affected. Overall, our findings indicate that future work may be able to train language-specific models more efficiently by focusing exclusively on the \textsc{Emb}, $G$, $W_\uparrow$ and \textsc{LMH} parameters.

\begin{table*}[t]
\centering
\small
  \begin{tabular}{lcccccccc|l}
    \toprule
    \textsc{Task} $\rightarrow$                  & \textsc{LA} & \textsc{NER} & \textsc{Senti} & \textsc{Summ}& \textsc{CSR} & \textsc{QA} & \textsc{TM} & \textsc{CT} & AVG. \\
    $\downarrow$ \textsc{Model}                  & ($mF_1$) & ($\mu F_1$) & ($mF_1$)  & (BERTS.) & (Acc.) & ($F_1$) & (Acc.) & (Acc.) & \\
    \midrule
        \multicolumn{10}{c}{\color{blue}{\textsc{\textbf{\llama{} based LLMs}}}} \\
    \midrule
    \llamabase{}                    & 33.43 & 22.31 & 61.54 & 65.50 & 29.76      & 63.54 & 38.69 & 57.05 & 46.48                 \\
    \llamachat{}                    & 47.42 & 24.63 & 62.35 & 66.15 & \cb{32.24} & 61.34 & 46.67 & 55.18 & 49.50                 \\
    \midrule
    \llamabase{} + \itda{}          & 36.10 & 28.48 & 62.86 & 66.43 & 29.04      & 64.40 & 49.10 & 58.46 & 49.35                 \\
    \llamachat{} + \itda{}          & 43.40 & 29.70 & 65.92 & 65.81 & 30.95      & 62.46 & 57.26 & 55.59 & 51.39                 \\
    \textsc{Viking-7B}              & 33.67 & 17.18 & 49.48 & 61.96 & 25.11      & 56.29 & 23.97 & 34.90 & 37.82                 \\
    \midrule
    \rowcolor{gray!7}
    \sprogmodel{}                   & \cb{56.28} & 19.91      & 57.42      & 58.95      & 30.47 & 18.52      & 69.14 & 60.93 & 46.45                 \\
    \rowcolor{gray!7}
    \snakmodel{}                    & 52.91      & \cb{29.76} & \cb{66.70} & \cb{66.61} & 29.46 & \cb{64.66} & \cb{71.05} & \cb{71.88} & \cb{56.63}\textsuperscript{\cb{$\uparrow$10.15}}                 \\
    \midrule
    \multicolumn{10}{c}{\color{orange}{\textbf{\textsc{Mistral-7B based LLMs}}}} \\
    \midrule
    \textsc{Mistral-7B-v0.1}                       & 38.38      & \co{32.66}    & \co{54.53} & 66.47      & 37.39       & \co{64.55}        & 64.50 & 71.56 & 53.76                 \\
    \textsc{Munin-7B-alpha}     & 53.03      & 28.71         & 43.77      & 67.27      & \co{42.68}  & 63.44             & 83.01 & 77.91 & 57.48                 \\
    \textsc{Munin-7B-v0.1dev0}  & \co{57.02} & 28.74         & 50.72      & \co{67.89} & 42.17       & 64.41             & \co{93.45} & \co{85.82} & \co{61.28}\textsuperscript{\co{$\uparrow$7.52}}                  \\
    \bottomrule
  \end{tabular}
    \caption{\textbf{Results (Test) on the ScandEval Benchmark.} We evaluate \llamabase{}, as well as the chat version against \snakmodel{} and other 7B models in ScandEval (best results in blue). In the subsequent rows, we test the same \llama{} tuned the Danish instruction tuning data (+ \itda{}). In the final rows, we show the Mistral-based models (best results in orange). We evaluate in $F_1$, macro-averaged $F_1$ ($mF_1$), micro-averaged $F_1$ ($\mu F_1$), BERTScore (BERTS.;~\citealp{Zhang2020BERTScore}), and Accuracy (Acc.).}
  \label{tab:scandeval-results}
\end{table*}

\section{Final Results and Analysis}\label{sec:results}

\paragraph{Benchmark Results.} Using our final model configurations, we present our results on the test split of the Danish portion of ScandEval in~\cref{tab:scandeval-results}. We compare \snakmodel{} against variants built on the same base model, including the original \llamabase{} and \llamachat{}. In addition, we train +\itda{} variants of these English \llama{} models on the same Danish instruction datasets as \snakmodel{}, in order to isolate the effect of Danish language modeling pre-training. Finally, we include comparisons to the Viking-7B model \citep{viking2024} and similarly-sized models based on the Mistral model suite \citep{jiang2023mistral,dfm2024munin}.

Overall, \snakmodel{} outperforms all other \llama{}-based models, including those with access to the same set of Danish instruction-tuning data, with a final average benchmark score of 56.63. The performance improvements over the English model are particularly pronounced for sub-tasks based on natural Danish data, including \textsc{LA} (33.43 $\rightarrow$ 52.91), \textsc{TM} (38.69 $\rightarrow$ 71.05), and \textsc{CT} (57.05 $\rightarrow$ 71.88). While the Mistral-7B-based models outperform \snakmodel{} by up to 4.65\% abs., this approximately matches the base model performance difference between Mistral-7B-v0.1 and \llamabase{} which spans 7.28\%. 
 
\paragraph{Qualitative Behaviors.} Since ScandEval scores are largely computed using constrained generation, we would like to highlight some qualitative observations from when models generate text without constraint. First, we find that \llama{} models fail to generate Danish text consistently, even when explicitly prompted to do so (confirming the findings by~\citealp{puccetti-etal-2024-ai}). Since they nonetheless achieve non-trivial benchmark scores under constrained generation, we hypothesize, that they obtain some Danish language functionality during their original, primarily English pre-training. Our custom \llama{} models to which we add Danish instruction tuning (+\itda{}) generate Danish responses (even when prompted in English), highlighting that a relatively small amount of translated Danish instructions is sufficient to bias models towards generating output in a new language. Nonetheless, the fact that \snakmodel{}, which is trained on non-translated Danish text outperforms the models trained on translated data, highlights the importance of curating high-quality native-language data for the adaptation target.

\section{Guidance for Future Work}\label{sec:guidance}
From our final evaluation, as well as our analysis of the training dynamics of \snakmodel{}, we next consolidate some guidance for future work adapting English LLMs to languages with similar linguistic properties and resource constraints.

\paragraph{Data.}
As we found large overlaps across data sources, as well as large amounts of non-Danish or irrelevant data (\cref{sec:data}), applying stringent pre-processing standards is important when working with smaller languages---especially when automatic filtering tools may be biased towards larger, related languages (e.g., Swedish).

\paragraph{Training.} Our training dynamics analysis (\cref{sec:training-dynamics}) showed that despite our total 13.6B word pre-training corpus, applying instruction tuning after 2,000--5,000 steps of Danish pre-training (i.e., less than half of the corpus) may already be sufficient to obtain close-to-final performance. For instruction tuning itself, one epoch over translated data appears to be sufficient to amplify instruction-following functionalities in the target language. Nonetheless, training on non-translated target language data is important to improve performance on more culturally specific tasks based on native data (i.e., LA, TM, and CT).

Finally, our weight divergence analysis revealed that most parameter updates are consolidated in the embeddings, feed-forward up-projections, and language modeling head. As English and Danish share a relatively similar syntactic structure, languages with more distinctive typologies may nonetheless exhibit larger changes to the self-attention parameters. For model adaption across a comparable typological distance as English and Danish however, focusing training efforts on the aforementioned parameter types---in addition to employing existing parameter-efficient fine-tuning techniques (e.g., \citealp{hu2022lora,dettmers2023qlora})---may therefore yield even higher efficiency gains.

\section{Conclusion}
In this work, we introduced the \textsc{SnakModel} suite, which includes a 7B-parameter base and instruction-tuned LLM for Danish, in addition to its pre-training and instruction-tuning data, intermediate checkpoints, and evaluation. 
By analyzing design decisions related to data curation and training dynamics, we further consolidated guidelines for future work adapting LLMs to new languages, to foster research not just in Danish, but in language communities with similar resource constraints.

\section*{Limitations}\label{app:limitations}

\paragraph{What Went Wrong and What Decisions We Took.}
Our training process encountered several challenges across multiple runs. In Run 1, we began by restarting training from the \llama{} checkpoint using the identical learning rate the original model had been trained on. However, we faced gradient explosion at iteration 2,031, which we attempted to mitigate through gradient clipping. Despite this effort, server crashes at step 3,500 and persistent gradient explosions forced us to halt the run after approximately 46 days, with a final language model loss of $\pm$1.77. For Run 2, we halved the peak learning rate to 1.5 $\times$ 10\textsuperscript{$-4$} and adjusted other parameters, but gradient explosion recurred at step 1,390, leading us to terminate the run after about 10 days with a final loss of $\pm$1.79. In Run 3, we significantly reduced the peak learning rate to 1.5 $\times$ 10\textsuperscript{$-5$}, reasoning that as we were continuing pre-training, we should aim for a rate lower than Llama2's final learning rate. This approach has shown effective, with the training reaching iteration 12,500 after approximately 93 days and achieving a language model loss of $\pm$1.72.

\section*{Acknowledgments}
First, we would like to thank Barbara Plank for allowing us to use her compute hardware for this period of time. Additionally, this work was impossible without the stable High Performance Compute cluster at the IT University of Copenhagen, being able to train a model for $\pm$90 days without a single interruption is extraordinary. Second, we thank Ahmet Üstun for giving us invaluable and concrete comments on hyperparameter setup for continuous pre-training. Last, we would also like to thank the reviewers for their valuable comments. Elisa Bassignana is supported by a research grant (VIL59826) from VILLUM FONDEN. Mike Zhang is supported by a research grant (VIL57392) from VILLUM FONDEN.

\bibliographystyle{acl_natbib}
\bibliography{nodalida2025, anthology}


\end{document}